# On The Construction of Extreme Learning Machine for Online and Offline One-Class Classification - An Expanded Toolbox

**Chandan Gautam[1]\*, Aruna Tiwari[1], Qian Leng[2]**

[1]**Indian Institute of Technology Indore**

[2]**Microsoft ATC**

E-mail: chandangautam31@gmail.com, artiwari@iiti.ac.in, leng_qian@qq.com

**Abstract:** One-Class Classification (OCC) has been prime concern for researchers and effectively employed in various disciplines. But, traditional methods based one-class classifiers are very time consuming due to its iterative process and various parameters tuning. In this paper, we present six OCC methods and their thirteen variants based on extreme learning machine (ELM) and Online Sequential ELM (OSELM). Our proposed classifiers mainly lie in two categories: reconstruction based and boundary based, where three proposed classifiers belong to reconstruction based and three belong to boundary based. We are presenting both types of learning viz., online and offline learning for OCC. Out of six methods, four are offline and remaining two are online methods. Out of four offline methods, two methods perform random feature mapping and two methods perform kernel feature mapping. We present a comprehensive discussion on these methods and their comparison to each other. Kernel feature mapping based approaches have been tested with RBF kernel and online version of one-class classifiers are tested with both types of nodes viz., additive and RBF. It is well known fact that threshold decision is a crucial factor in case of OCC, so, three different threshold deciding criteria have been employed so far and analyses the effectiveness of one threshold deciding criteria over another. Further, these methods are tested on two artificial datasets to check there boundary construction capability and on eight benchmark datasets from different discipline to evaluate the performance of the classifiers. Our proposed classifiers exhibit better performance compared to ten traditional one-class classifiers and ELM based two one-class classifiers. Through proposed one-class classifiers, we intend to expand the functionality of the most used toolbox for OCC i.e. DD toolbox. All of our methods are totally compatible with all the present features of the toolbox.

**Keywords:** One-class classification (OCC), Extreme Learning Machine (ELM), Online Sequential ELM (OSELM), One-Class ELM (OCELM), Autoassociative ELM (AAELM)

1. Introduction

Novelty or outlier detection [1] has been always prime attention of researchers in various disciplines and one-class classifier [2, 3] has been broadly applied for this purpose. One-class classification (OCC) was coined by Moya et al. [4]. It becomes necessity when data of only one class is available or the data belongs to other classes is very rare. For example, if we want to classify between healthy and unhealthy people. In this case, it is possible to define the range of the data for the healthy people but not possible to define such a definite range for unhealthy people. Because it is possible to prepare the data based on only existing diseases but not possible to prepare based on forthcoming diseases. Thus the data which is available for training can be taken as positive or normal samples and rest of the samples are not possible to define or very rare or unknown, can be taken as negative or abnormal. These abnormal samples are termed as outliers. Sometimes, it is difficult to have outliers when training samples are available. The reason behind the absence of outlier samples can be the very high measurement costs or the less occurrence of an event, i.e. failure of nuclear power plant, a rare medical disease [5] or machine fault detection [6] etc. Even if outlier examples are available, there is a possibility that those examples are not well distributed or do not characterize all possible issues of outliers, so, we can't rely on the outlier examples. Therefore, one-class classification has been broadly applied in the field of novelty detection [1, 7, 8]. Overall goal of one-class classifier is to accept target samples and reject the outliers. Basic assumption of multi class classification is also followed here that the examples belong to same class share same pattern among them. So, it can be applied in various type of problems [2] viz., (i) Novelty detection (ii) Outlier detection (iii) Classification of imbalanced data.

Various methods have been proposed to resolve the one-class classification problem. According to Pimental et al. [1], these methods can be broadly divided into five categories: (i) Probabilistic or density based (ii) Distance based (iii) Information theoretic techniques (iv) Domain or boundary based (v) Reconstruction based. Tax [2] divided OCC methods in three parts viz., density based, boundary based and reconstruction based. We have expanded the toolbox of OCC provided by Tax [2], so we divided our proposed work based on the category provided by Tax [2] only. We will provide detailed discussion about this in the next section. This paper is mainly focused on the last two categories i.e. domain or boundary and reconstruction based. Our literature survey also primarily focuses on these two categories only.

The remaining paper is organized as follows. Literature survey about OCC is discussed in Section 2. Section 3 presents motivation of our proposed work. Section 4 provides a brief description of ELM and OSELM. Section 5 discusses about proposed work and three threshold deciding criteria. Subsequently, Section 6 highlights the difference between proposed classifiers and other existing classifiers. Section 7 describes experimental design; Section 8 presents result and discussion. At last, Section 9 evinces conclusion and future direction of work.

## 2. Literature Survey

OCC word is firstly coined by Moya et al. [3] and applied for target recognition application. As far as learning for OCC is concerned, it can be divided into three parts [9] based on the data applied during learning (i) with positive examples only (ii) with positive and small amount of negative examples only (iii) with positive and unlabeled data. Ritter et al. [10] applied OCC for outlier detection and automatic chromosome classification. Japkowicz [11] proposed autoassociative based approach for OCC in the absence of counterexamples. OCC have been widely studied by Tax [2] and he developed various models to handle OCC problem viz., three models are density based viz., (i) Gaussian model (ii) mixture of Gaussians and (iii) Parzen density estimator, three models are boundary based viz., (i) k-centers method (ii) KNN-dd and (iii) SVDD and four models are reconstruction based viz., (i) k-mean clustering (ii) self-organizing maps (iii) PCA and mixtures of PCA's and (iv) diabolo networks. Tax and Duin [12] and Scholkopf [13] proposed algorithm to handle OCC problem based on support vector machine (SVM) using positive examples only. Manevitz and Yousef [14, 15] employed OCSVM and autoassociative neural network based approach for retrieving the interested document from the internet in presence of positive examples only. In text classification community, learning with positive and unlabeled data [16, 17] have received much attention. Koppel et al. [18] applied OCC for authorship verification problem. Wang and Stoflo [19] deployed Naïve Bayes for masquerade detection task. One-class k-nearest neighbor (KNN) based approach has been applied by Munroe and Madden [20] for vehicle model recognition from images and showed that it provides comparable outcomes to that of multi class classifiers. One-class classification based on Random Forest [21] and based on Firefly algorithm [22] has also been presented in recent years. After surveying various papers, it has been observed that OCSVM has been explored more compared to other one-class methods. So, there are various variants of OCSVM has been proposed like OCSVM based on hidden information [23], Covariance-guided OCSVM [24], ensembling of OCSVM [25], Multi Task Learning by OCSVM [26], Fuzzy OCSVM [27] etc. Most recently, Leng et al. [28] proposed OCC based on Extreme Learning Machine (ELM) where ELM consists of only one node in output layer. Leng et al. [28] tested their model with only one type of threshold deciding criteria i.e. rejection of few percentages of most deviant training samples after completion of training on all samples. Ravi and Singh [29] have also proposed voting based single class classifier using autoassociative extreme learning factory but didn't compare with any existing one-class classifier and they have used only random based feature mapping. ELM has also been applied for anomaly or intrusion detection by using binary classification [30, 31].

## 3. Motivation Of Our Work

A neural network can be explored with four possible variations as you can see in **Fig. 1**. Our proposed methods are based on ELM and explored it for OCC. Lot of work [32, 33, 34] have been carried out which exhibited the superiority of ELM over traditional machine learning techniques like Back-propagation (BP), SVM, Probabilistic Neural Network (PNN), Multilayer Perceptron (MLP) etc. in terms of generalization capability and training time for regression, binary and multiclass classification. Traditional methods need to tune the weights and various parameters in each iteration but ELM provides result just in one pass. Since, existing OCC models are based on traditional methods, so, problem lies with traditional methods also remain with existing OCC models. As an example, backpropagation based OCC requires many iterations to stabilize the weight, so it will be very time consuming compared to single hidden layer feed forward network (SLFN). We are going to propose six methods with their thirteen variants of OCC based on ELM and Online Sequential Extreme Learning Machine (OSELM) [35] (this paper is an extension of the paper presented in ELM-2015 [36]). Due to the fast growing need of streaming data, online version of OCC is required. Earlier, ELM and OSELM had been presented for binary, multiclass classification and regression tasks (see **Fig. 1**) but we are presenting it for one-class classification. It will enrich one-class classification with various novel features of OSELM as mentioned below [35]:

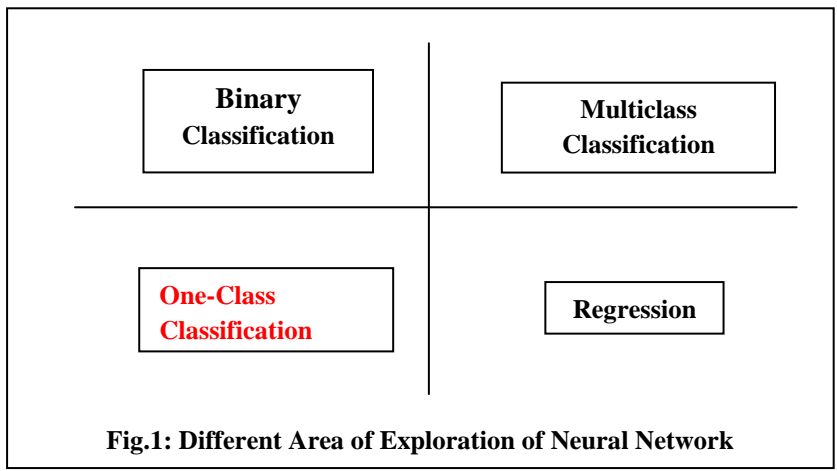

**Fig.1: Different Area of Exploration of Neural Network**

(i) Training samples can be presented in sequential manner i.e. either one by one or chunk by chunk.
(ii) A chunk or single sample will be removed from the network as soon as the training procedure of that particular sample will be completed; it has no need to wait for other chunk or observation.
(iii) Whenever new data will arrive in training set, there is no need to train whole data again, just train that particular new data.
(iv) No prior knowledge about the size of the training data is required.

## 4. Preliminaries & Overview Of ELM And Their Variants

ELM is proposed by Huang et al. [33, 34] for addressing the slow learning speed of traditional neural network. Before moving to proposed algorithm, we will briefly discuss about basic ELM. Following notations are being used throughout the paper:

**X**= Training set $\{X_1, X_2......X_N\}$, *where* $X_i = \{x_{i1}, x_{i2},...., x_{in}\}$, where, i=1, 2…N. Here, 'N' denotes number of samples and 'n' denotes number of features in input layer.

**Y**= Testing set $\{Y_1, Y_2......Y_M\}$, *where* $Y_i = \{y_{i1}, y_{i2},...., y_{in}\}$, where 'M' denotes number of samples and 'n' denotes number of features in input layer.

**T**= Output layer matrix.

**m** = Number of hidden nodes.

**Kernel_name**= {Random, RBF, Linear, Polynomial or Wavelet}.

**Thr**= Out of three threshold deciding criteria {Thr1, Thr2, Thr3} (it is mentioned in the further section), select which threshold criteria would be employed for outlier detection.

**Thresh**= Determined threshold value during training for distinguishing between outlier and target data.

**W**=$[w_{i1}, w_{i2}......, w_{in}]$ is the input weight vector connecting from the j$^{th}$ feature (input) to the i$^{th}$ hidden neuron, where, i=1, 2….m, which is determined during training.

**b** = Bias for i$^{th}$ hidden node $[b_1, b_2......, b_m]$.

**g(_)** = Activation Function

**β** = $[\beta_{i1}, \beta_{i2}......, \beta_{ik}]$ is the output weight vector connecting the i$^{th}$ hidden neuron and the k$^{th}$ output neurons. Where i=1, 2…m, which is determined during training.

**E** = $\{e_1, e_2......e_i\} \in \Re$, where $e_i$ is the outlier score of corresponding samples from X or Y.

**O** = $\{o_1, o_2......o_i\} \in \Re$, where $o_i$ is the predicted output of corresponding samples from X or Y.

**f** = number of folds only use to generate validation set from training data.

**sigma_thr**= Threshold require for decision boundary during model selection.

**fracrej**= Fraction of rejection require from target sample for model selection.

**N₀**: Number of initial training data used in the initial phase of **OS-OCELM** or **OS-AAELM**, where $N_0 >= m$ and $N_0 <= N$.

**Block/Chunk** = Size of block of data learnt by **OS-OCELM** or **OS-AAELM** in each step.

**Param_in**= Various input parameters which are required to that specific one-class classifier (like {kernel parameter (**kern_par or σ**) and regularization coefficient **C** i.e. [**C, kern_par**] for basic ELM based one-class classifier} and {[**m, N₀, block**] for OSELM based one-class classifier}).

### 4.1 Brief Overview of ELM

As proposed by Huang et al. [33, 34], we can state basic ELM in three steps. For N given training samples, activation function g(x) and by assuming m number of hidden neurons following steps are taken to perform learning of the neurons,

1. Random initialization of input weights $W_i$ and bias $b_i$ for all m number of nodes/neurons, where i=1, 2....m.
2. Calculate the hidden layer output matrix H by applying X over all m number of hidden neurons with W and b with activation function g as $g(WX+b)$.
3. Calculate the output weight $\beta = H^\dagger T$, Where, $H^\dagger = (H^T H)^{-1} H^T$ and T=Output layer. Here, Objective of ELM is to minimize the training error as well as norm of the output weights [33,34] in the following eqnuation (1):

   Minimize: $\|H\beta - T\|^2$ and $\|\beta\|$, (1)

Where,

$$H = \begin{bmatrix} g(W_1 X_1 + b_1) & \cdots & g(W_m X_1 + b_m) \\ \cdots & \cdots & \cdots \\ g(W_1 X_N + b_1) & \cdots & g(W_m X_N + b_m) \end{bmatrix}_{N \times m}$$

$$\beta = \begin{bmatrix} \beta_1^T \\ \cdot \\ \cdot \\ \cdot \\ \cdot \\ \cdot \\ \beta_m^T \end{bmatrix}_{m \times k} \quad T = \begin{bmatrix} T_1 \\ \cdot \\ \cdot \\ \cdot \\ \cdot \\ \cdot \\ T_N \end{bmatrix}_{N \times k}$$

## 4.2 Brief Overview of OSELM

OSELM [35] is originated by the concept of batch learning of ELM. Training samples can be presented in sequential manner i.e. either one by one or chunk by chunk with fixed or varying length and remove those observation from training whenever training would be completed. It consists of 2 phase:

**Initialization phase:**

1. Initially let us set k=0, where k denotes $k^{th}$ chunk of data.
2. Training is initialized by small chunk of training data. $N_0$ is the number of initial training data within a chunk, which should not be less than the number of hidden neurons. Initialize the random hidden nodes parameter, input weights $w_i$ and bias $b_i$, $i = 1 \cdots m$ same as basic ELM, where m represents number of hidden nodes.
3. Calculate the initial hidden layer output matrix $H_0 = [g(W_i, b_i, X_j)]$, where i=1 to m and j=1 to $N_0$, and g is an activation function.
4. Calculate initial weight matrix between hidden and output layer, $\beta_0 = P_0 H_0^T T_0$, where $P_0 = (H_0^T H_0)^{-1}$ and $T_0$ denotes corresponding class level of training data.

**Sequential learning phase:**

For $(k+1)^{th}$ chunk of new observation, $N_{k+1}$ denotes number of training data in the $(k+1)^{th}$ chunk.

(a) Calculate the partial hidden layer output matrix $H_{k+1}$ by using same Moore-Penrose inverse formula used in ELM.
(b) Calculate the output weight matrix $\beta^{(k+1)}$ as in equation (2) [35] as follows:

$$\beta^{(k+1)} = \beta^{(k)} + P_{k+1} H_{k+1}^T (T_{k+1} - H_{k+1} \beta^k) \quad (2)$$

Where, $P_{k+1} = P_k - P_k H_{k+1}^T (I + H_{k+1} P_k H_{k+1}^T)^{-1} H_{k+1} P_k$.

## 4.3 Brief Review of other variants of ELM

In recent years, ELM has been well expanded in both dimensions: theory and application. Various variants have been proposed: incremental ELM [37, 38], semi-supervised and unsupervised ELM [39], circular-ELM [40], two stage ELM [41], parsimonious ELM [42], Forgetting Online Sequential ELM [43], weighted ELM [44], multi-layer ELM [45, 46], parallel ELM [47, 48, 49, 50], hardware implementation of ELM [51, 52], domain adaption transfer learning based ELM [53, 54], local receptive fields based ELM [55] etc. ELMs have built some connection between machine learning techniques and biological learning mechanisms [56, 57], so it attracted researcher towards it. Recently, Anton et al. [58] has presented a high performance ELM toolbox for big data application. We are also presenting a toolbox for one-class classification using ELM in this paper. A thorough surveys on ELM has been done by Huang et al. [59] which includes many more variants and related work on ELM.

## 5. Proposed Classifiers And Their Different Threshold Criteria

Earlier, Data Description (DD) Toolbox for MATLAB has been presented by Tax [60] for OCC. This toolbox is most usable toolbox for OCC since many years and it is also enriched with various other facilities like visualization of classifier etc. So, we incorporated our proposed methods with this toolbox in such a way so that, the inbuilt features of toolbox are fully compatible with our proposed methods. Our proposed methods are mainly based on two types of OCC viz., reconstruction based one-class ELM i.e. autoassociative ELM (**AAELM**) and boundary based one-class ELM (**OCELM**). The Kernelized version of both **AAELM** and **OCELM** viz., autoassociative Kernelized ELM (**AAKELM**) and one-class Kernelized ELM (**OCKELM**), is also presented in this paper. In all methods, only positive samples are used for training. Now onwards, positive samples will be called as target data and negative samples as outliers. Rest of this section is as follows: proposed classifiers are discussed in Section 5.1 and their various threshold criteria are discussed in Section 5.2. For model selection, we employed consistency based model selection [61] that is discussed in Section 5.3.

**5.1 Proposed classifiers: An Expanded Toolbox for One-Class Classification**

Our proposed methods can be divided in two broad categories (i) Boundary based and (ii) reconstruction based as depicted in **Fig. 2**.

**Assumption for boundary based one-class classifier:** "Model is trained by only target data X and endeavored to approximate all data to one because training have only one class. Since, weights between layers are trained according to pattern of target data i.e. positive data. But if any pattern other than target data will be provided to trained model then it will not be properly approximate to one. Therefore, difference between approximated value and one will be large if pattern is differing from target data. If difference will be more than the threshold then treat that sample as outlier otherwise belongs to target data."

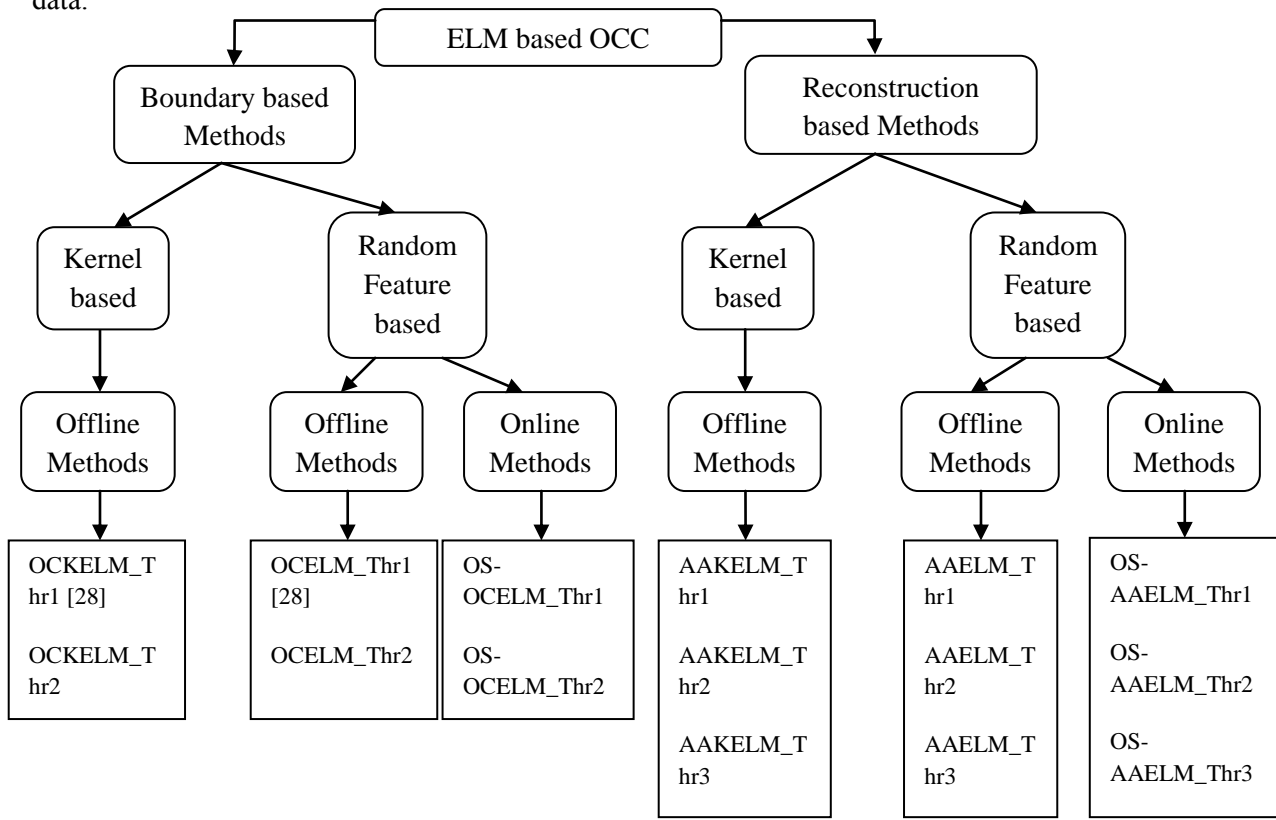

**Fig. 2: All ELM based One-Class Classification Methods**

**Assumption for reconstruction based one-class classifier:** "Reconstruct the input vector X at the output layer during training and weights between layers are trained only for target data i.e. positive data. So, if any pattern other than target data is passed to the model then that pattern will not be well reconstructed. Therefore, difference between input and output in case of outlier will be more than the difference between input and output of target dataset. So, we can decide threshold in such a way that if dissimilarity is more than this threshold then assume as outlier data otherwise target data".

**5.1.1 One-Class Extreme Learning Machine using randomized (OCELM) and kernelized feature mapping (OCKELM):**

**OCELM** is boundary based One-Class Extreme Learning Machine using randomized feature mapping (i.e. Random Kernel with Sigmoid activation function) and **OCKELM** is also boundary based One-Class Kernelized Extreme Learning Machine using kernelized feature mapping. Both of the methods work with two of three different threshold deciding criteria (**Thr1** or **Thr2**). **Thr3** can't be applied on boundary based OCC (reason is explained in Section 5.2). Our proposed Classifiers (i.e. OCELM and OCKELM) are different from the classifier proposed by Leng et al. [28] in term of threshold selection criteria as we introduce one more threshold criteria for OCELM and OCKELM. Threshold selection is the most crucial part of OCC because it decides whether any data point belongs to that classifier or not. So, we have employed two different thresholds deciding criteria to this classifier and generated four different variants viz., **OCELM_Thr1 [28], OCELM_Thr2, OCKELM_Thr1 [28] and OCKELM_Thr2**. Among four variants, two variants viz., **OCELM_Thr1 and OCKELM_Thr1** had been proposed by Leng et al. and two novel variants viz., **OCELM_Thr2 and OCKELM_Thr2** are proposed in this paper. Architecture of **OCELM** is described in **Fig. 3**.

**Principles:** It follows above discussed assumption of the boundary based one-class classifier. Apart of that, the first and second layer of proposed classifier is similar to basic ELM but output layer contains only one node as it belongs to only one class. Here, input layer contains data from only one class, so all data belong to same feature space and therefore we set target of each sample to any real number R (i.e. R=1 in our case). It works with both random and kernel feature mapping.

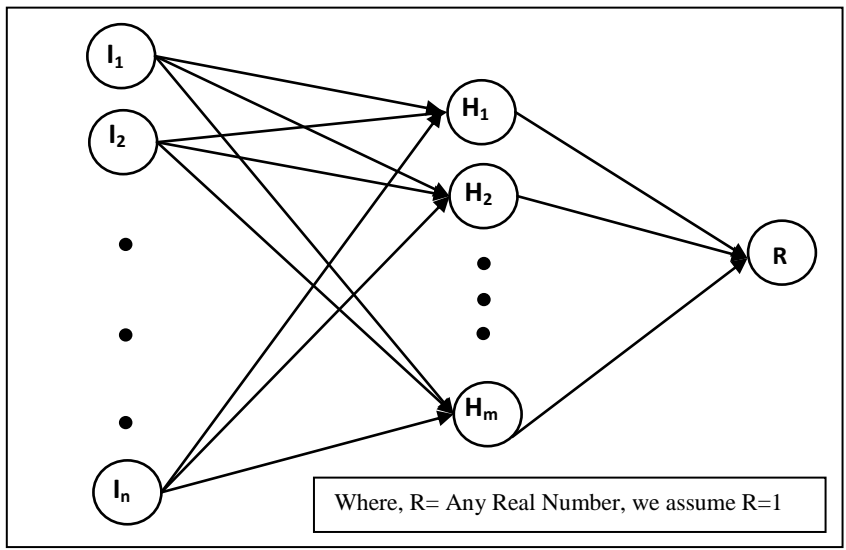

**Fig. 3: Architecture of One-Class Extreme Learning Machine (OCELM)**

Algorithm for this classifier is mentioned below:

---
**Algorithm for OCELM and OCKELM**

**Function OCELM** (If random feature selection)/ **OCKELM** (If kernel feature selection)
**Input:** X, T, kern_par, Kernel_name, Thr, fracrej
**Output:** Thresh, **β**

    if (Kernel_name == 'Random')  then     // Random feature selection is selected
      // Calculate the hidden layer output matrix as following
      $H = g(W\ X\ + b)$     //Where g - activation function, W- input weights and b - bias
      H= H * transpose (H)     // Create Random Kernel
      kern_par= [W, b]
    else     // Kernel feature selection is selected
      H= Kernel_name (kern_par, X)
    end if
**Set target at any real number R for all samples, we assume R=1 i.e. T=[1, 1….1].**
//Calculate the output weight
**β =** ((H + speye (N) / C) \ (**T**))     // Where, speye (N) = Sparse identity matrix of size N-by-N.
**O= H * β**     // Compute output O=[$o_1, o_2……o_N$]
if (Thr==Thr1) then     **// Refer Section 5.2 for Threshold Deciding Criteria**
    for i=1 to N
      **$e_i$ = absolute ($o_i$ – R)**     // Where $\mathbf{E} = \{e_1, e_2……e_i\}$ and i=1, 2….N.
      **Here, R can be any real number and we assume R=1**
    end
    sorted_error= sort(**E**)     // Sort the error in descending order
    index= round(fracrej*N)
    Thresh= sorted_error(index)
else (Thr==Thr2)
    for i=1 to N
      **$e_i$ = (($o_i$ – R) * ($o_i$ – R))**     // Where $\mathbf{E} = \{e_1, e_2……e_i\}$ and i=1, 2….N.
    end
    Thresh= mean (**E**) + 0.2 * Standard deviation (**E**).
end if
**end Function**

---

### 5.1.2 Autoassociative Extreme Learning Machine using randomized (AAELM) and kernelized feature mapping (AAKELM):

AAELM and AAKELM are reconstruction based one-class classifier. **AAELM** uses randomized feature mapping (i.e. Random Kernel with Sigmoid activation function) and **AAKELM** uses kernelized feature mapping with anyone of the three different threshold deciding criteria as discussed in Section 5.2 (**Thr1, Thr2** or **Thr3**). AAELM architecture has been used earlier also but for different purpose like clustering [62], data imputation [63], deep learning [64] etc. Here, we present **AAELM** and **AAKELM** for one-class classification. Architecture of **AAELM** is described in **Fig. 4**. The first and second layer of the proposed classifiers is similar to basic ELM but output layer and input layer contain same number of nodes as it is autoassociative in nature.

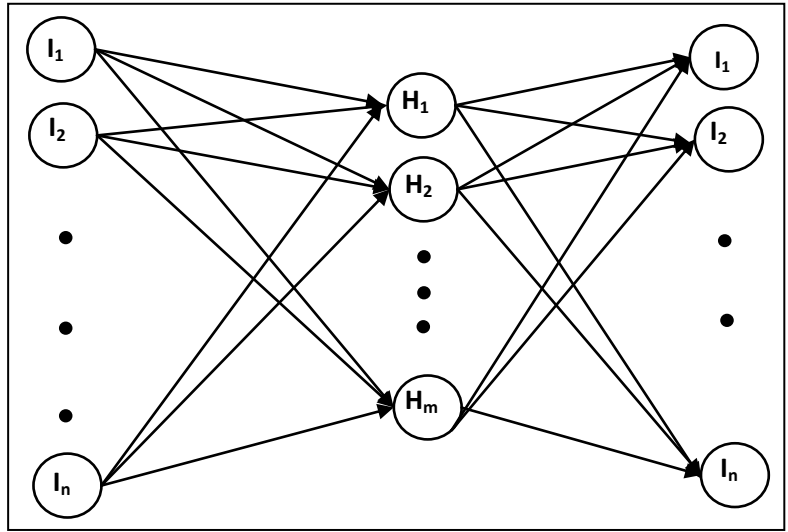

**Fig. 4: Architecture of Autoassociative Extreme Learning Machine (AAELM)**

Here, we are reconstructing input data at output layer and calculate difference between actual and predicted as error. As we know that threshold deciding criteria is a crucial part in **OCC** task. So, we have employed three different thresholds deciding criteria to this architecture and generated six novel variants for **OCC** task viz., **AAELM_Thr1, AAELM_Thr2, AAELM_Thr3, AAKELM_Thr1, AAKELM_Thr2** and **AAKELM_Thr3**.

**Principles:** It follows above discussed assumption of the reconstruction based one-class classifier. Apart of that, the first and second layer of the proposed classifiers is similar to basic ELM but output layer contains identical node as input node because it is an autoassociative method. So, we decide outlier based on the fact that whether input is properly reconstructed or not. It works with both random and kernel feature mapping.

Algorithm for this classifier is mentioned below:

**Algorithm for AAELM and AAKELM**

**Function AAELM** (If random feature selection)/ **AAKELM** (If kernel feature selection)
**Input:** X, T, kern_par, Kernel_name, Thr, fracrej
**Output:** Thresh, **β**

    if (Kernel_name == 'Random') then    // Random feature selection is selected
      // Calculate the hidden layer output matrix as following

      $H = g(WX + b)$    //Where g - activation function, W- input weights W and b - bias

      H= H * transpose (H)    // Create Random Kernel
      kern_par= [W, b]
    else    // Kernel feature selection is selected
      H= Kernel_name (kern_par, X)
    end if
    **Set target as input for all samples i.e. T= X**
    //Calculate the output weight
    **β =** ((H + speye (N) / C) \ (T))    // Where, speye (N) = Sparse identity matrix of size N-by-N.

```
        O= H * β                              // Compute output O=[o₁, o₂……oₙ]
     if (Thr==Thr1) then                      // Refer Section 5.2 for Threshold Deciding Criteria
        for i=1 to N
```
$$e_i = \sum_{j=1}^{n}(x_{ij} - o_{ij})^2$$   // Where $E = \{e_1, e_2......e_i\}$, i=1, 2….N and n= number of features
```
        end
        sorted_error= sort(E)        // Sort the error in descending order
        index= round (fracrej*N)
        Thresh= sorted_error(index)
     else if (Thr==Thr2)
        for i=1 to N
```
$$e_i = \sum_{j=1}^{n}(x_{ij} - o_{ij})^2$$   // Where $E = \{e_1, e_2......e_i\}$, i=1, 2….N and n= number of features
```
        end
        Thresh= mean (E) + 0.2 * Standard deviation (E).
     else
        As threshold criteria Thr3 is explained in Section 5.2
     end if
end Function
```

### 5.1.3 Online Sequential One-Class Extreme Learning Machine (OS-OCELM) Using Randomized Feature Mapping:

It is boundary based Online Sequential One-Class Extreme Learning Machine (**OS-OCELM**) using randomized feature (i.e. Random Kernel with **Sigmoid** and **RBF** activation function) with two different threshold deciding criteria (**Thr1** or **Thr2**). **Thr3** can't be applied on boundary based OCC as discussed in Section 5.1. So, two different threshold deciding criteria has been employed to the **OS-OCELM** and generated two novel variants for OCC task viz., **OS-OCELM_Thr1 and OS-OCELM_Thr2.** We experimented our model with both types of hidden nodes viz., additive and hidden.

**Principles:** It follows same principle as mentioned in Section 5.1.1 with one enhancement i.e. it can learn from data sequentially. Online sequential learning saves a lot of time if your training data is continuously growing, because there is no need to train data again from scratch, just train on newly arrived data. So, we enabled our proposed classifier **OCELM** for online sequential learning. It works with only random feature mapping.

Output weight calculation formula of OSELM i.e. equation (2) will be modified for **OS-OCELM** as equation (3) i.e. Set Target as any real number R (where, we assume R=1) due to one-class classification.

$$\beta^{(k+1)} = \beta^{(k)} + P_{k+1}H_{k+1}^T(R - H_{k+1}\beta^k), \text{ where } P_{k+1} = P_k - P_k H_{k+1}^T(I + H_{k+1}P_k H_{k+1}^T)^{-1}H_{k+1}P_k \quad (3)$$

**Steps required for this classifier are mentioned below:**
1. Select type of hidden nodes (additive or RBF) and their corresponding activation function g.
2. Phase I: Initialization Phase

// Random feature mapping is selected
(a) For additive hidden nodes:
   Assign random input weight W and bias b
   For RBF hidden nodes:
   Centre W and impact factor b
(b) Calculate hidden layer output matrix (H0) for initial training data of size N0 as:
   $\mathbf{H_0} = g(\mathbf{W}, \mathbf{b}, \mathbf{X_0})_{N_0 \times m}$, where $X_0 = X_j$, where j =1 to $N_0$, and g is an activation function.
(c) Calculate initial output weight matrix $\boldsymbol{\beta_0} = \mathbf{P_0}\mathbf{H_0^T}R$, where $\mathbf{P_0} = (\mathbf{H_0^T}\mathbf{H_0})^{-1}$ and **R denotes target value, which can be any real number, here we assume R=1** i.e. class column is set to 1 for all data points.
(d) Set k=0.

3. Phase II: Sequential Learning Phase
   While (any data chunk remains to process)
   For $(k+1)^{th}$ chunk of new observation, $N_{k+1}$ denotes number of observations in the $(k+1)^{th}$ chunk.
   (a) Calculate the partial hidden layer output matrix $H_{k+1}$ by using same Moore-Penrose inverse formula used in ELM.
   **(b) Set Target as any real number R (where, we assume R=1) due to one-class classification.**
   (c) Calculate the output weight matrix $\beta^{(k+1)}$ :
   (d) $\boldsymbol{\beta^{(k+1)}} = \boldsymbol{\beta^{(k)}} + \mathbf{P_{k+1}}\mathbf{H_{k+1}^T}(R - \mathbf{H_{k+1}}\boldsymbol{\beta^k})$,
   Where, $\mathbf{P_{k+1}} = \mathbf{P_k} - \mathbf{P_k}\mathbf{H_{k+1}^T}(\mathbf{I} + \mathbf{H_{k+1}}\mathbf{P_k}\mathbf{H_{k+1}^T})^{-1}\mathbf{H_{k+1}}\mathbf{P_k}$.
   (e) Set k=k+1.
   End of For
   End of While
   So, calculation of output weight is finished and final $\beta$ is calculated.

4. Calculate the hidden layer output matrix as following:
   $H = g(W, b, X)$, where g is an activation function either Sigmoid or RBF.
5. Compute output **O= H * β**
6. Apply two of the three threshold criteria viz., **Thr1** or **Thr2** (as mentioned in Section 5.2) similar as described in Section 5.1.1 with **OCELM**.

### 5.1.4 Online Sequential Autoassociative Extreme Learning Machine (OS-AAELM) Using Randomized Feature Mapping::

It is reconstruction based one-class classifier i.e. Online Sequential **AAELM** (**OS-AAELM**) using randomized feature (i.e. Random Kernel with **Sigmoid** or **RBF** activation function) with three different threshold deciding criteria (**Thr1, Thr2** or **Thr3**). So, three different threshold deciding criteria have been employed to the **OS-AAELM** and generated three novel variants for **OCC** task viz., **OS-AAELM_Thr1, OS-AAELM_Thr2 and OS-AAELM_Thr3.** Our model is experimented with both type of hidden nodes viz., additive and RBF.

**Principle:** It follows the same principle as discussed in Section 5.1.2 with one enhancement i.e. it can learn from data sequentially. As we have discussed that online sequential learning saves lot of time if training data is growing over time, because, there is no need to train the data again from scratch, just train on newly arrived data. So, we enabled our proposed classifier **AAELM** for online sequential learning. It works with only random feature mapping.

Output weight calculation formula of OSELM i.e. equation (2) will be modified for **OS-AAELM** as equation (4) i.e. set input X as target due to autoassociative nature so that T=X:

$$\boldsymbol{\beta}^{(k+1)} = \boldsymbol{\beta}^{(k)} + \mathbf{P}_{k+1}\mathbf{H}_{k+1}^{T}(\mathbf{X}_{k+1} - \mathbf{H}_{k+1}\boldsymbol{\beta}^{k}), \text{ where, } \mathbf{P}_{k+1} = \mathbf{P}_{k} - \mathbf{P}_{k}\mathbf{H}_{k+1}^{T}(\mathbf{I} + \mathbf{H}_{k+1}\mathbf{P}_{k}\mathbf{H}_{k+1}^{T})^{-1}\mathbf{H}_{k+1}\mathbf{P}_{k} \quad (4)$$

**Steps required for this classifier are mentioned below:**
1. Select type of hidden nodes (additive or RBF) and their corresponding activation function g.
2. Phase I: Initialization Phase
   // Random feature mapping is selected
   (a) For additive hidden nodes:
       Assign random input weight W and bias b
       For RBF hidden nodes:
       Centre W and impact factor b
   (b) Calculate hidden layer output matrix (H0) for initial training data of size N0 as:
       $\mathbf{H}_0 = \mathbf{g}(\mathbf{W}, \mathbf{b}, \mathbf{X}_0)_{N_0 \times m}$, where $X_0 = X_j$, where j = 1 to $N_0$, and g is an activation function.
   (c) Calculate initial output weight matrix $\boldsymbol{\beta}_0 = \mathbf{P}_0\mathbf{H}_0^T\mathbf{X}_0$, where $\mathbf{P}_0 = (\mathbf{H}_0^T\mathbf{H}_0)^{-1}$ and $\mathbf{X}_0$ **denotes initial chunk of data as mentioned in previous step (b)**.
   (d) Set k=0.
3. Phase II: Sequential Learning Phase
   While (any data chunk remains to process)
       For (k+1)$^{th}$ chunk of new observation, $N_{k+1}$ denotes number of observations in the (k+1)$^{th}$ chunk.
       (a) Calculate the partial hidden layer output matrix $H_{k+1}$ by using same Moore-Penrose inverse formula used in ELM.
       **(b) Set input X as target due to autoassociative nature i.e. T=X.**
       (c) Calculate the output weight matrix $\beta^{(k+1)}$:
       (d) $\boldsymbol{\beta}^{(k+1)} = \boldsymbol{\beta}^{(k)} + \mathbf{P}_{k+1}\mathbf{H}_{k+1}^{T}(\mathbf{X}_{k+1} - \mathbf{H}_{k+1}\boldsymbol{\beta}^{k})$,
       Where, $\mathbf{P}_{k+1} = \mathbf{P}_{k} - \mathbf{P}_{k}\mathbf{H}_{k+1}^{T}(\mathbf{I} + \mathbf{H}_{k+1}\mathbf{P}_{k}\mathbf{H}_{k+1}^{T})^{-1}\mathbf{H}_{k+1}\mathbf{P}_{k}$.
       (e) Set k=k+1.
       End of For
   End of While
   So, calculation of output weight is finished and final $\beta$ is calculated.
4. Calculate the hidden layer output matrix as following:
   $H = g(W, b, X)$, where g is an activation function either Sigmoid or RBF.
5. Compute output **O = H * β**

6. Apply any one of the three threshold criteria viz., **Thr1, Thr2** or **Thr3** (as mentioned in Section 5.2) similar as described in Section 5.1.2 with **AAELM**.

## 5.2 Threshold Deciding Criteria for the Proposed Methods

In one-class classification, threshold plays crucial role in distinguishing outlier from target data. Three different criteria are considered here for determining threshold. One point must be noted that only positive samples are used for deciding the threshold criteria.

1) **Thr1:** Calculate the error using Euclidean distance between actual and predicted on each training data and arrange the error in decreasing order. Afterwards, set the threshold at rejection of 10% most erroneous data i.e. false negative rate should be expected at the rate of 10%.
2) **Thr2:** Calculate the error between actual and predicted on training and set the threshold (Thr) using following formula:
$$Thr2 = Error + 0.2 * Std$$
Where, Error = Mean square error over all the training data
Std = Standard deviation of error
3) **Thr3:** This threshold criteria is only for reconstruction based methods i.e. not for boundary based methods because here, we calculate difference between actual and predicted value of each feature and boundary based methods have only one node at output layer. Calculate relative error between actual and predicted on each attribute of training dataset by using modified relative error formula with reconstruction based one-class classifier. The following formula is generally used for relative error calculation:

$$err_i = abs\left(\frac{act_i - pred_i}{act_i}\right), \text{ where } i=1, 2 \ldots \ldots n.$$

$act_i$ = Actual Value and $pred_i$ = Predicted Value

**Remark:** The above formula will not work in the case when actual value will be zero. So, we modified relative error formula by adding predicted value in denominator as follows:

$$err_i = abs\left(\frac{act_i - pred_i}{act_i + pred_i}\right),$$
where i=1, 2 …….n.

```
Well_reconstructed_attribute=0
for i=1 to n
if err_i< thr_condn1 then
Well_reconstructed_attribute=
Well_reconstructed_attribute +1;
end if
end for
```

This error is a deciding factor that whether attribute is well reconstructed in output layer or not. There are 2 conditions require to satisfy for this threshold deciding criteria. We need to satisfy first condition (**condn1**) at this point for deciding how many features are well reconstructed at output layer. Afterwards, we need another condition (**condn2**) for deciding whether data should be treated as genuine or outlier. If number of `Not_Well_reconstructed_feature` is above **condn2** then it will be treated as target data otherwise outlier.

We considered **condn1**=0.5 i.e. if feature is generated with more than 50% error then it will be considered as not well reconstructed data otherwise `Well_reconstructed_feature`. We considered **condn2**= 0.1 * n i.e. if more than 90% attributes are well reconstructed then consider that sample as target sample otherwise outlier sample.

**Remark:** Among all three threshold criteria, **Thr1** and **Thr2** rejected few percent of data for deciding threshold but **Thr3** doesn't reject any percentage of data for deciding the threshold. We will see impact of this fact on artificial dataset in Results and Discussion section.

### 5.3 Consistency-Based model selection for the proposed one-class classifiers

For selecting model with optimized parameter, we performed consistency based model selection [61] with our proposed methods. This model selection works based on 2-sigma bound around error of classification model. It determines a threshold error (err_thr) by using equation (5) and run our proposed classifier with different parameters of classifier till it yields less error than err_thr.

**err_thr = (M*fracrej + sigma_thr*sqrt (fracrej*(1-fracrej)*M))/M**

$$= fracrej + sigma\_thr*sqrt\ (fracrej*(1-fracrej)/M). \quad (5)$$

where M=(N/f), where f= number of folds and sigma_thr= Threshold require for determining decision boundary during model selection.
Where,
**M** denotes number of sample in validation set.
**M*fracrej** is the expected number of rejected objects.
**fracrej*(1-fracrej)*M** is the variance and after square root of this is standard deviation.
**(M*fracrej + sigma_thr*sqrt(fracrej*(1-fracrej)*M))** is the maximum allowed number of rejected target objects.

This model selects parameter by linear search and when error of validation set is below of threshold error (**err_thr**) then stop searching further. If more than one parameter require for classifier then it will search for optimal parameter on all possible combination over range of the parameters. So, we need to provide the proper range of parameters.

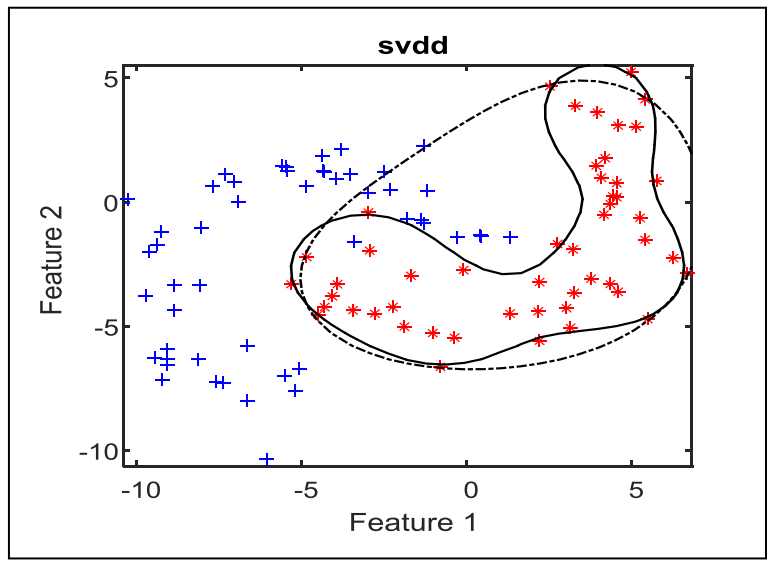

**Fig. 5: svdd OCC with Optimal and Without Optimal Parameter**

**Reason behind using consistency based model selection:** We selected optimized model for all presented methods (proposed and existed) in this paper. If we will not select optimal parameter then classifier may not be able to create proper boundary. As you can see in **Fig. 5**, support vector data description (**svdd**) created boundary around banana-shaped dataset with optimal (i.e. by using consistency based model selection) and without optimal parameter. Thick line boundary denotes with optimal parameter and dotted line boundary denotes without optimal parameter. It is clearly visible that dotted line also created boundary around banana-shaped dataset and covered all the data points but boundary is not optimal. So, it covers few data points of other classes also. Therefore, consistency based model selection is required for OCC. Following steps are required for consistency based model selection:

1) Decide the 2 sigma bound for finding most complex classifier as following:
   err_thr= fracrej + sigma_thr*sqrt (fracrej*(1-fracrej)/M), where M= (N/f), where f= number of folds. Number of folds and sigma_thr are passed by user during function call.
2) Divide the training dataset into **f** folds i.e. dividing into train and validation set.
3) For all possible parameter combination (which is selected from Param_in), check that for which parameter, our proposed classifier yields less average error over f folds than err_thr calculated in Step 1. Then pass the same parameter as optimized parameter for that classifier in remaining runs. So, in that way, it optimized the parameter as well as consumes less time.

## 6. How Our Proposed One-Class Classifier Is Different From Traditional OCC

Various classifiers have been proposed by various researchers and applied in various disciplines. But Tax [2] has done resounding work on OCC. Tax [2] proposed various classifiers as we discussed in literature. Tax also provided a MATLAB toolbox viz., DD toolbox [60] to employ these classifiers. We employed following classifiers with various datasets for comparison purpose:

a) Density based methods: (i) Parzen density data description (parzen_dd) (ii) Naive Parzen density data description (nparzen_dd) (iii) Gaussian data description (gauss_dd)
b) Boundary based methods: (ii) K-nearest neighbor data description (knndd) (ii) Minimax probability machine data description (mpm_dd) (iii) Support vector data description (svdd) (iv) Incremental svdd (incsvdd)
c) Reconstruction based methods: (i) Auto-encoder neural network data description (autoenc_dd) (ii) k-means data description (kmeans_dd) (iii) Principal component data description (pca_dd) (iv) Self-Organizing Map data description (som_dd)

These one-class classifiers have been implemented using DD Toolbox [60]. For optimal performance of these methods, we employed consistency-based model selection for one-class classification [61] using DD Toolbox [60]. These are all offline classifiers. We propose an expansion of DD toolbox for both online and offline OCC based on ELM and OSELM. As we have discussed, existing autoencoder is a time consumable method during training because it needs to tune the weight properly but our all three proposed autoencoders based methods have no need to tune the weight and yield the result in just one pass. Our proposed methods can be utilized according to the requirement as mentioned here: (i) AAELM – when random feature mapping is required (ii) AAKELM – when kernelized feature mapping is required, it is useful in those cases where feature mapping is not known, (iii) OS-AAELM – It also uses random feature mapping but an online sequential version of AAELM. It will be useful for those cases where online learning is required. All existing one-class classifiers, which use activation function, that can work only for differentiable function but our all proposed methods in the expanded toolbox can work for both differentiable and non-differentiable activation function.

## 7. Experimental Setup and Dataset Description

### 7.1 Dataset Description and Its Preprocessing

Performance of the proposed methods has been tested on variety of datasets as mentioned in **Table 1**. We tested our proposed classifiers on two artificial and eight benchmark datasets. Artificial datasets are created by the help of PRToolBox [65] and each dataset is of size hundred. Those datasets are banana-shaped and ring-shaped dataset. **Banana-shaped dataset** is of two dimensional two class dataset with a banana shaped distribution. The data is uniformly distributed along the bananas and is superimposed with a normal distribution with standard deviation 1 in all directions. Second, **Ring-shaped dataset** is one-class 2D circular dataset with radius 1 superimposed with 1D normally distributed radial noise with standard deviation 0.1. Benchmark datasets are downloaded from UCI Machine Learning Repository [66] and website of TU Delft [67]. We prepared the dataset for OCC in the same way as prepared by Leng et al. [28]. Perform z-score on dataset such that attributes of the dataset are centered to have mean 0 and scaled to have standard deviation 1. For sample data $x$ with mean $\overline{X}$ and standard deviation $S$, the z-score of a data point $x$ is $\left(\frac{x-\overline{X}}{S}\right)$. Divide the dataset according to their classes. Assume normal samples as target class and remaining of the classes as outlier classes. Follow the same procedure for each dataset. **Fig. 6** provides the overall flow of our classification task. Further, divide the target class data in two equal parts – one part is used for training the model and other part is combined with outlier class for testing. 50% data from target class is selected randomly before performing any classifier on the dataset and remaining 50% of target data is merged with outlier data for testing. Now, training have only target class and testing have both target (target class is also called as positive class) and outlier (outlier is also called as negative class) class. We have experimented over twenty runs and the calculate average of output over all runs as final output. In each run, we randomly select 50% target data for training and remaining all data for testing but we don't calculate optimal parameters in each run; however training and testing set get change in each run due to random selection of samples in each run. We just calculate optimal parameter in first run only, so it speeds up the execution and saves a lot of time. Apply trained model for testing set Y and obtained outlier score or error E and verify following condition for outlier detection (here, Thresh value is obtained during training of classifier): If E < Thresh then target class otherwise an outlier. Calculate average performance using performance evaluation criteria like AUC, F1 etc. over twenty runs.

**Table 1: Dataset Description**

| Dataset Name | Number of Attributes | Number of Targets | Number of Outliers | Name of Target Class |
|---|---|---|---|---|
| Abalone | 10 | 1407 | 2770 | Classes 1-8 |
| Arrhythmia | 278 | 237 | 183 | Normal |
| Breast Cancer | 9 | 458 | 241 | Benign |
| Diabetese | 8 | 500 | 268 | Present |
| Ecoli | 7 | 52 | 284 | Periplasm |
| Liver | 6 | 145 | 200 | Healthy |
| Sonar | 60 | 111 | 97 | Mines |
| Spectf | 44 | 95 | 254 | 0 |

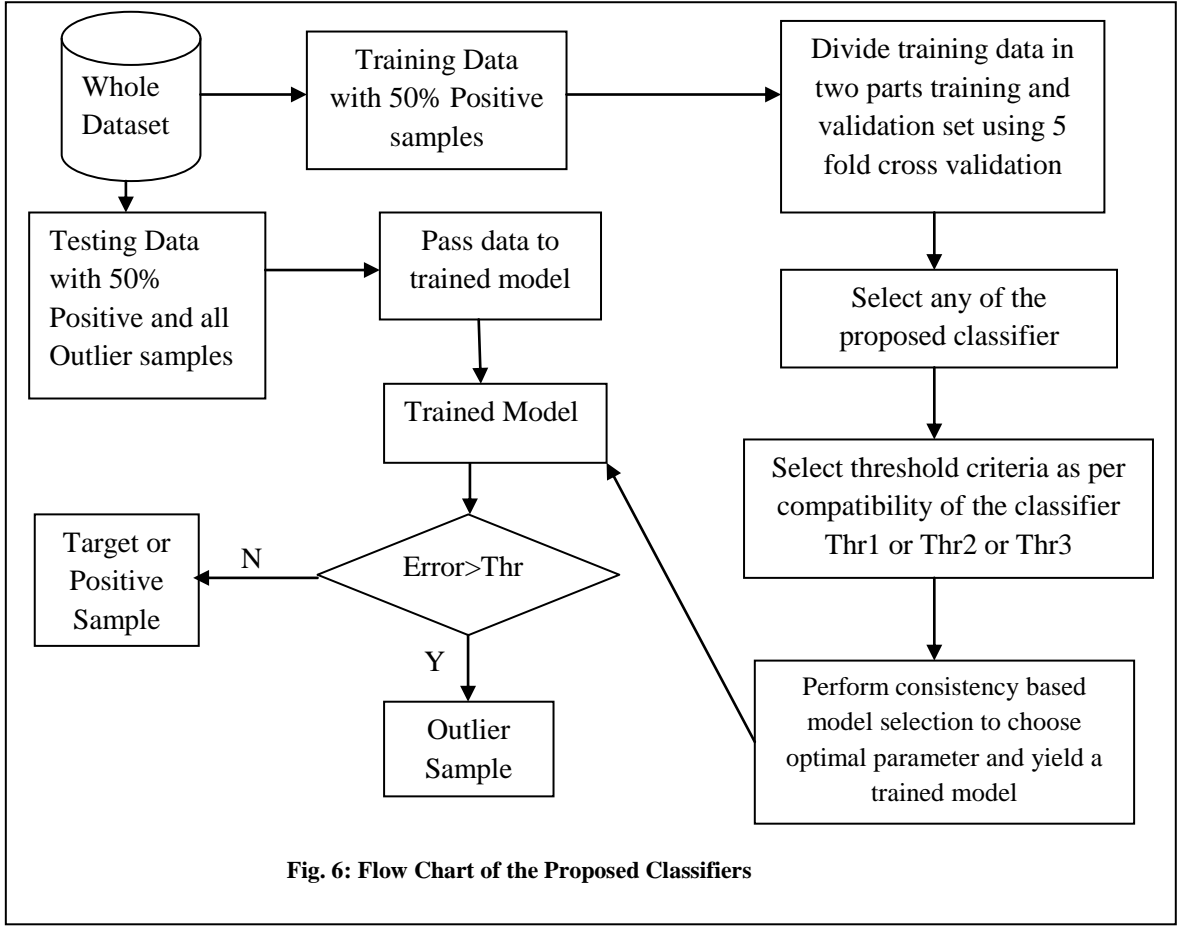

**Fig. 6: Flow Chart of the Proposed Classifiers**

**7.2 Performance Measures**

All experiments have been executed on MATLAB 2011a in Windows 7 (64 bit) environment with 4 GB RAM, 3.10 GHz processor and Intel i5 processor. We first preprocessed the data and normalize in the range [0 1] then the proposed approaches are applied for each training/testing break up. We executed each classifier twenty (write textual representation for numbers wherever necessary) times and calculate the average of twenty runs for each performance evaluation measure. For the purpose of evaluation of performance of the classifier, we calculated various measures viz., precision, recall (Sensitivity), F1 [68], specificity, accuracy and Area Under Curve (AUC) [69]. Since, accuracy alone can't provide the total insight about the quality of classification; therefore we calculated F1 and AUC also. F1 is a measure to combine Recall and Precision i.e. harmonic mean of Recall and Precision. F1 is also providing biased insight because it is made up of Recall and Precision only, which is also called as true positive rate and positive predictive value respectively. Therefore, F1 and accuracy alone do not always provide true picture of classifier, therefore we also calculated AUC. AUC considers both sensitivity and specificity in calculation so it provides better insight regarding classification. We will provide F1, Accuracy and AUC in our classification results but considered superiority of one classification methods over other by using AUC. Above mentioned classification evaluation criteria are defined as follows:

(a) Precision (P) = TP / (TP+FP)

(b) Recall (R) or Sensitivity = TP / (TP+FN)
(c) Specificity (SP) = TN / (TN+FP)
(d) F1 = (2 * Precision * Recall) / (Precision + Recall)
(e) Accuracy (Acc) = (TP + TN) / (TP + TN + FP + FN)
(f) Area Under Curve (AUC) = ½(Sensitivity + Specificity)

We experimented three of our proposed methods viz., **AAELM, AAKELM, OS-AAELM** with three different threshold criteria **Thr1**, **Thr2** and **Thr3** & six of our proposed methods viz., **OCELM, OCKELM, OS-OCELM, AAELM, AAKELM, OS-AAELM** with only two different threshold criteria **Thr1** and **Thr2** (detailed discussion in the next section). Basic ELM based approaches have been tested with sigmoid activation function and our kernelized based approaches are tested on RBF kernel by optimal selection of two different parameters viz., regularization (**C**) and kernel parameter (**kern_par**). However, our expanded toolbox can handle four kernel functions viz., RBF, linear, polynomial and wavelet. **OSELM** based OCC approaches have been tested with both additive and RBF hidden nodes. Impact of these parameters and kernel on our proposed methods are discussed in the next section.

## 8. Results and Discussion

The proposed methods are tested on two artificial datasets and eight benchmark datasets. Their performances are evaluated by optimally selected numerous parameters viz., number of hidden neurons, kernel parameters, regularization parameter. We experimented with three threshold deciding criteria viz., **Thr1, Thr2** and **Thr3**. By employing **Thr1** and **Thr2** on all six proposed methods, we generated two existing [28] and ten novel variants of ELM for OCC, (i) boundary based methods viz., **OCELM_Thr1 [28], OCELM_Thr2, OCKELM_Thr1 [28], OCKELM_Thr2, OS-OCELM_Thr1, OS-OCELM_Thr2,** (ii) reconstruction based methods viz., **AAELM_Thr1, AAELM_Thr2, AAKELM_Thr1, AAKELM_Thr2, OS-AAELM_Thr1, OS-AAELM_Thr2**. By employing **Thr3** with **AAELM**, **AAKELM and OS-AAELM**, we proposed three novel variants of ELM for OCC i.e. reconstruction based methods viz., **AAELM_Thr3, AAKELM_Thr3, OS-AAELM_Thr3**. Following discussion will provide the performance of our proposed methods on artificial and benchmark datasets.

### 8.1 Artificial Datasets

Two artificial datasets viz., banana-shaped and ring-shaped datasets, have been employed for testing the proposed method's ability on boundary creation. **Banana shaped dataset** is to evaluate the convexity of the proposed methods. **Fig. 7** exhibits the behavior of the proposed methods on creation of convex decision boundary around banana-shaped dataset. Most of the methods are able to create proper decision boundary but some methods are unable to create lower convex part of banana. Second artificial dataset, **Ring-shaped dataset** is to test the ability of classifier that how they can create internal boundary of the ring because that boundary differentiate it from only circular boundary. Boundary based proposed methods created boundary better than reconstruction based methods except one **AAKELM_Thr3** for both datasets. Although, boundary of **AAKELM_Thr3** is not smooth but it covers all points of dataset wherever remaining methods lack in covering all points. Reason behind that its threshold deciding criteria **Thr3** does not reject any fraction of data during training however other threshold deciding criteria do. But, **Thr3** does not work well with other random feature selection based proposed methods as you can realize with **AAELM_Thr3** and **OS-AAELM_Thr3** for banana-shaped data in **Fig. 7(j), (s) and (t)** & they do not show any boundary in our plot for ring dataset, therefore we didn't present diagram for **AAELM_Thr3** and **OS-AAELM_Thr3** in case of ring dataset. We tested **AAKELM_Thr3** with other artificial datasets to check its ability and it created boundary successfully as well as covered all points.

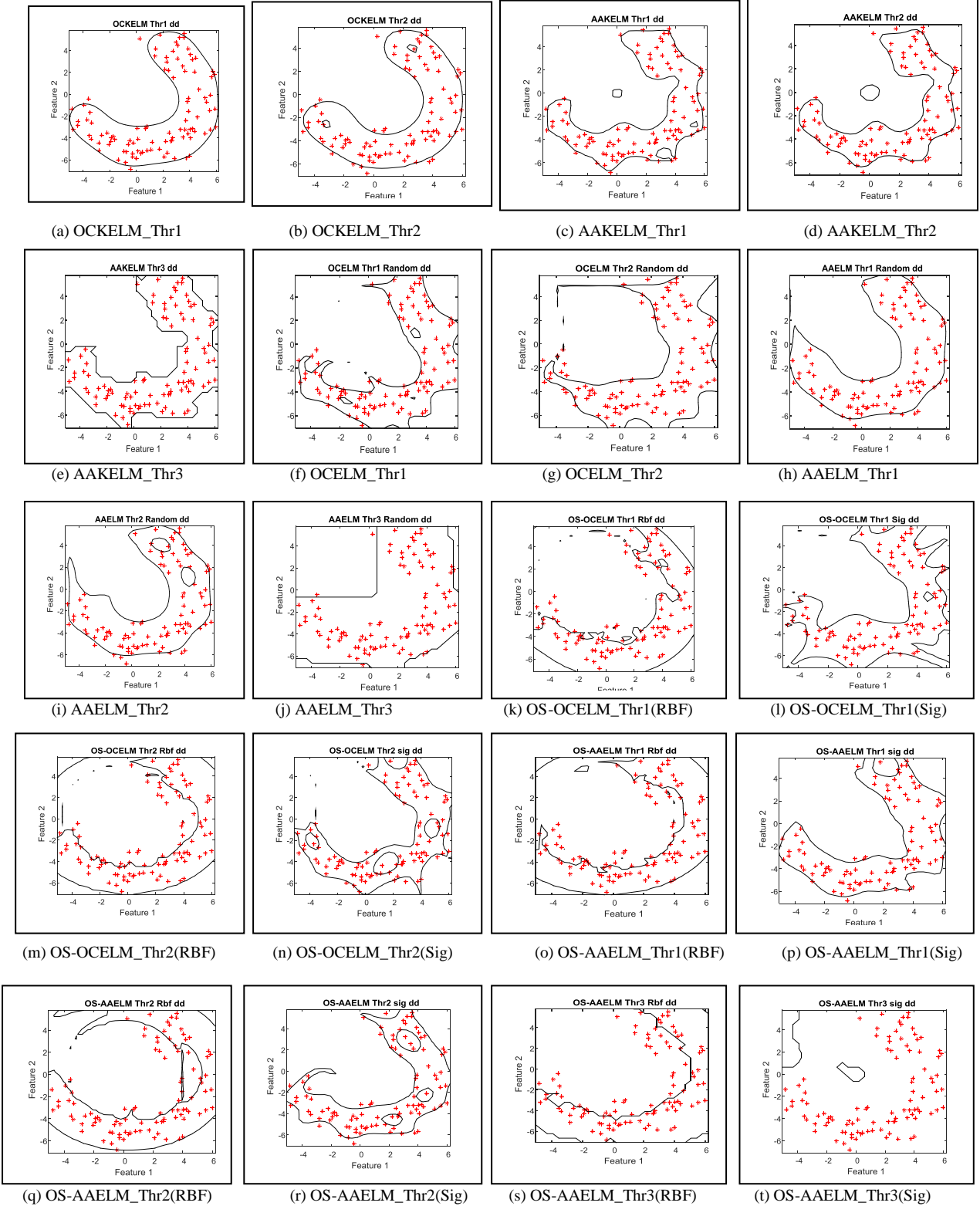

**Fig. 7(a)-(t): Construction of Boundary for Banana Shaped Dataset by Various Methods**

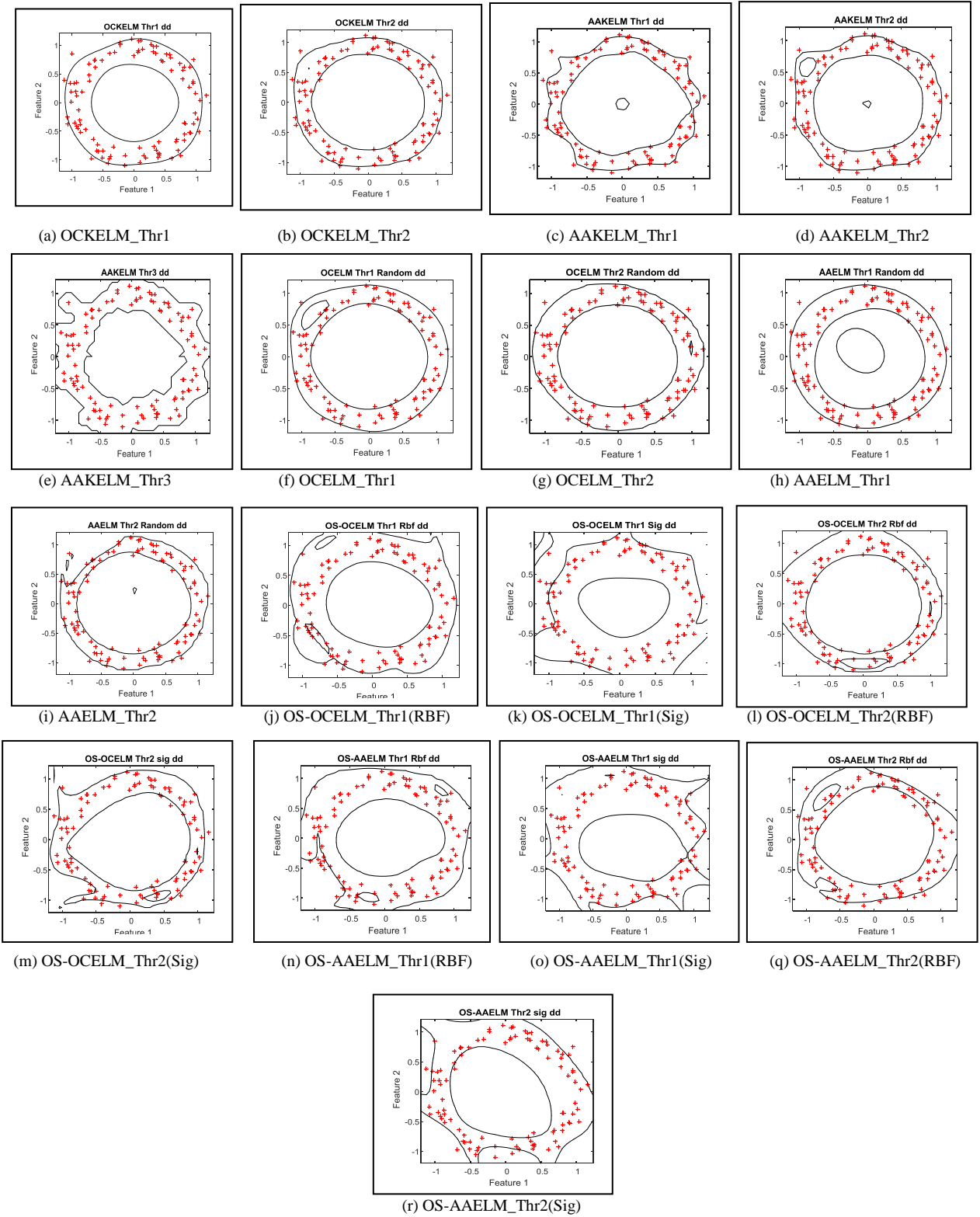

**Fig. 8(a)-(t): Construction of Boundary for Ring Shaped Dataset by Various Methods**

Online sequential version of OCC methods does not create proper boundary, they shows underfitting and also covered extra spaces as you can see in **Fig. 8**. Reason behind that online sequential version is not regularized and also not kernelized however all offline version considered regularization factor in account also. Our future work will be based on inclusion of both of the factors in online sequential version of one-class classifiers.

**8.2 Benchmark Datasets**

As we have realized with artificial dataset that kernel based methods are better in boundary construction, same has been experienced with real life datasets. Kernel based methods performed better than random feature based methods for most of the dataset as you can see in **Table 2(a)-(d)**. We will compare the performance of various methods based on their AUC value because ACC can show biased behaviour as we have discussed in earlier section. However for sake of comparison from the result reported by Leng et al. [28], we have calculated F1 for all the methods reported by Leng et al. [28] and achieved approximately same result as reported in paper [28]. We have also presented standard deviation of AUC over twenty runs.

**In Abalone dataset,** best ACC and AUC are not achieved by same methods but as we have discussed already that AUC will be deciding criteria. So, **AAKELM_Thr1** outperformed rest of the methods in term of AUC and also exhibited least standard deviation among all the methods. So, reconstruction based methods viz., **AAKELM_Thr1**, **AAELM_Thr1** and **OS-AAELM_Thr1** performed best among kernel feature based offline methods, random feature based offline methods and random feature based online methods respectively. **In Arrhythmia dataset**, **AAELM_Thr1** performed best among all the proposed and existing methods. All methods with **Thr3** resulted in NAN (i.e. Not a Number) because out of twenty runs if any run have value NAN then there average will also lead to NAN. NAN occurs actually in Precision calculation because all true positive and false negative leads to zero, so, Precision finally lead to NAN. One point to be noted here that, random feature mapping based method performs better than kernel feature mapping based methods among ELM based OCC methods. **In Breast cancer dataset**, nearest neighbor based OCC method performed best among all methods but **AAKELM_Thr3** exhibited comparable result with only difference of 0.3848. The surprise performance of **knndd** due to local learning capability of **knndd** because nearest neighbor based method performs well only when data is locally binds very well. **In Diabetes dataset**, **AAKELM_Thr2** and **OS-AAELM_Thr1** performed very similar and outperformed rest of the existing and proposed methods. **In Ecoli dataset**, again **knndd** performed best among all methods and **AAELM_Thr1** performed best among ELM based OCC methods. But one point to be noted, **knndd** performs better in term of AUC but **OCKELM_Thr2** performed better in term of ACC. But, ACC is providing biased picture regarding performance because sensitivity is 75.57% (i.e. low) and specificity is 98.38% (i.e. very high) of **OCKELM_Thr2**. AUC of **knndd** provides right picture because it provides balanced performance with sensitivity 89.03% and specificity 90.70%. **In Liver dataset**, we experienced same behavior of ACC and AUC i.e. **AAELM_Thr2** performed best in term of AUC and **mpm_dd** performed best in term of ACC. Moreover, sensitivity of **mpm_dd** is 15.69% (i.e. very low) only but specificity is 93.32% (i.e. very high). One more thing, random feature based method i.e. **AAELM_Thr2** performed better than kernel feature based methods. We experienced same as Ecoli and Liver dataset in case of **Sonar dataset** i.e. **OS-AAELM_Thr3** performed better in term of AUC and **AAKELM_Thr3** performed better in term of ACC. Moreover, **Thr3** based OCC methods performed better than other two threshold based methods.

Similar to above three datasets, **AAKELM_Thr3** performed better in term of AUC and **parzen_dd** performed better in term of ACC for **Spectf dataset**. Here also, **Thr3** based method i.e. **AAKELM_Thr3** outperformed all other methods.

**Table 2(a): Performance of all classifiers on Abalone and Arhythima datasets over average of 20 runs**

|  | Abalone | | | | Arrhythmia | | | |
|---|---|---|---|---|---|---|---|---|
|  | **F1** | **ACC** | **AUC** | **Std_AUC** | **F1** | **ACC** | **AUC** | **Std_AUC** |
| **Knndd** | 47.53 | 59.62 | 71.01 | 1.34 | 62.33 | 56.98 | 62.99 | 0.32 |
| **Svdd** | 54.38 | 77.77 | 73.14 | 1.04 | NAN | 60.80 | 50.00 | 0.00 |
| **kmeans_dd** | 45.09 | 56.07 | 68.36 | 1.23 | 66.46 | 66.46 | 69.75 | 1.78 |
| **parzen_dd** | 49.95 | 63.43 | 73.40 | 0.32 | 56.32 | 39.20 | 50.00 | 0.00 |
| **nparzen_dd** | 50.30 | 64.08 | 73.67 | 0.20 | 56.32 | 39.20 | 50.00 | 0.00 |
| **pca_dd** | 46.32 | 57.88 | 69.75 | 0.94 | **67.97** | 68.97 | **71.70** | 2.14 |
| **mpm_dd** | 49.24 | 70.72 | 70.48 | 0.99 | NAN | 60.80 | 50.00 | 0.00 |
| **Incsvdd** | 45.95 | 57.37 | 69.36 | 0.70 | 66.34 | 64.78 | 69.01 | 1.04 |
| **som_dd** | 41.07 | 47.22 | 63.23 | 3.96 | 66.93 | 66.46 | 70.06 | 1.19 |
| **autoenc_dd** | 48.50 | 61.12 | 71.59 | 4.60 | 65.87 | 65.08 | 68.79 | 1.56 |
| **gauss_dd** | 49.79 | 63.50 | 73.16 | 0.28 | NAN | 60.80 | 50.00 | 0.00 |
| **OCELM_Thr1** | 51.91 | 71.59 | 73.11 | 1.13 | 62.69 | 60.17 | 64.65 | 2.46 |
| **OCKELM_Thr1** | 52.68 | 68.04 | 75.41 | 0.68 | 67.46 | 68.01 | 70.97 | 1.08 |
| **Proposed Methods** | | | | | | | | |
| **OCKELM_Thr2** | 54.33 | 71.60 | 76.01 | 0.68 | 65.28 | 67.59 | 69.43 | 1.72 |
| **AAKELM_Thr1** | 58.57 | 76.50 | **78.57** | 0.35 | 59.15 | 58.21 | 61.61 | 1.38 |
| **AAKELM_Thr2** | 54.02 | 70.21 | 76.25 | 0.49 | 59.27 | 64.04 | 64.55 | 1.20 |
| **AAKELM_Thr3** | NAN | 79.76 | 50.00 | 0.00 | NAN | 60.80 | 50.00 | 0.00 |
| **OCELM_Thr2** | 52.68 | 74.86 | 72.71 | 1.12 | 62.98 | 63.46 | 66.30 | 2.94 |
| **AAELM_Thr1** | 59.24 | 78.67 | 77.87 | 0.73 | **67.28** | **70.27** | **71.67** | 1.37 |
| **AAELM_Thr2** | 59.81 | 80.24 | 77.40 | 0.95 | 62.50 | 68.36 | 68.26 | 2.32 |
| **AAELM_Thr3** | NAN | 79.76 | 50.00 | 0.00 | NAN | 60.80 | 50.00 | 0.00 |
| **OS-OCELM_Thr1(RBF)** | 55.56 | 75.78 | 75.36 | 1.14 | 57.10 | 50.38 | 56.34 | 4.54 |
| **OS-OCELM_Thr2(RBF)** | 54.61 | 75.19 | 74.60 | 0.92 | 57.32 | 54.29 | 58.51 | 4.67 |
| **OS-AAELM_Thr1(RBF)** | 59.57 | 79.52 | 77.57 | 1.48 | 58.38 | 53.21 | 58.60 | 3.95 |
| **OS-AAELM_Thr2(RBF)** | 60.26 | 80.66 | 77.40 | 1.11 | 56.24 | 56.84 | 59.29 | 5.65 |
| **OS-AAELM_Thr3(RBF)** | **62.47** | **86.50** | 75.28 | 3.37 | NAN | 60.80 | 50.00 | 0.00 |
| **OS-OCELM_Thr1(Sig)** | 52.97 | 73.22 | 73.67 | 0.94 | 58.08 | 59.20 | 61.59 | 2.81 |
| **OS-OCELM_Thr2(Sig)** | 50.45 | 70.95 | 71.71 | 1.02 | 56.48 | 59.49 | 60.96 | 3.91 |
| **OS-AAELM_Thr1(Sig)** | 57.16 | 77.42 | 76.26 | 0.75 | 59.23 | 59.62 | 62.36 | 4.32 |
| **OS-AAELM_Thr2(Sig)** | 55.52 | 76.22 | 75.04 | 0.88 | 55.64 | 55.13 | 59.13 | 4.04 |
| **OS-AAELM_Thr3(Sig)** | 40.46 | 41.56 | 61.10 | 7.02 | NAN | 60.80 | 50.00 | 0.00 |

**Table 2(b): Performance of all classifiers on Breast Cancer and Diabetese datasets over average of 20 runs**

|  | Breast Cancer | | | | Diabetese | | | |
|---|---|---|---|---|---|---|---|---|
|  | F1 | ACC | AUC | Std_AUC | F1 | ACC | AUC | Std_AUC |
| **knndd** | **95.12** | **95.48** | **95.36** | 1.17 | 67.90 | 58.75 | 59.82 | 0.82 |
| **svdd** | 79.57 | 83.60 | 83.17 | 3.88 | NAN | 51.74 | 50.00 | 0.00 |
| **kmeans_dd** | 93.11 | 93.76 | 93.59 | 2.08 | 68.48 | 60.72 | 61.65 | 1.55 |
| **parzen_dd** | 94.01 | 94.51 | 94.37 | 1.64 | 51.64 | 60.47 | 59.94 | 2.13 |
| **nparzen_dd** | 92.15 | 92.60 | 92.51 | 1.38 | 69.76 | 63.57 | 64.36 | 0.89 |
| **pca_dd** | 87.93 | 88.18 | 88.19 | 1.30 | 65.72 | 55.12 | 56.25 | 4.43 |
| **mpm_dd** | 71.59 | 78.46 | 77.89 | 1.22 | 20.95 | 56.11 | 54.63 | 1.17 |
| **incsvdd** | 94.18 | 94.39 | 94.36 | 1.14 | 66.51 | 55.98 | 57.15 | 1.43 |
| **som_dd** | 89.20 | 90.54 | 90.29 | 1.93 | 67.37 | 63.79 | 64.26 | 1.39 |
| **autoenc_dd** | 92.54 | 93.20 | 93.05 | 1.52 | 66.27 | 57.48 | 58.46 | 2.54 |
| **gauss_dd** | 94.85 | 95.23 | 95.11 | 1.19 | 66.34 | 55.98 | 57.12 | 0.54 |
| **OCELM_Thr1** | 79.50 | 80.20 | 80.17 | 2.98 | 63.07 | 50.51 | 51.76 | 1.30 |
| **OCKELM_Thr1** | 93.22 | 93.84 | 93.68 | 1.64 | 68.86 | 63.38 | 64.07 | 1.28 |
| **Proposed Methods** |  |  |  |  |  |  |  |  |
| **OCKELM_Thr2** | 91.34 | 92.27 | 92.06 | 1.73 | 68.48 | 62.39 | 63.14 | 1.04 |
| **AAKELM_Thr1** | 93.53 | 94.10 | 93.94 | 1.41 | 67.39 | 60.52 | 61.33 | 0.89 |
| **AAKELM_Thr2** | 92.81 | 93.48 | 93.31 | 1.33 | 68.05 | 65.42 | **65.78** | 1.51 |
| **AAKELM_Thr3** | 94.70 | 95.11 | 94.98 | 0.83 | 59.01 | 59.48 | 59.51 | 1.58 |
| **OCELM_Thr2** | 78.48 | 79.96 | 79.84 | 3.23 | 63.06 | 50.02 | 51.31 | 1.46 |
| **AAELM_Thr1** | 92.96 | 93.59 | 93.44 | 2.15 | **69.91** | 63.63 | 64.43 | 0.94 |
| **AAELM_Thr2** | 92.53 | 93.20 | 93.04 | 1.30 | 69.38 | 61.80 | 62.73 | 2.55 |
| **AAELM_Thr3** | 67.40 | 53.12 | 54.27 | 1.62 | 55.31 | 52.61 | 52.88 | 0.97 |
| **OS-OCELM_Thr1(RBF)** | 86.96 | 87.73 | 87.64 | 2.21 | 61.66 | 59.16 | 59.46 | 2.45 |
| **OS-OCELM_Thr2(RBF)** | 85.73 | 87.07 | 86.90 | 2.17 | 62.59 | 58.12 | 58.61 | 3.49 |
| **OS-AAELM_Thr1(RBF)** | 90.41 | 91.49 | 91.27 | 1.47 | 63.46 | 63.23 | 63.34 | 1.62 |
| **OS-AAELM_Thr2(RBF)** | 88.69 | 90.12 | 89.86 | 1.87 | 64.74 | 62.79 | 63.07 | 3.09 |
| **OS-AAELM_Thr3(RBF)** | 89.96 | 89.55 | 89.69 | 3.14 | 60.19 | 55.72 | 56.22 | 2.76 |
| **OS-OCELM_Thr1(Sig)** | 87.54 | 88.39 | 88.28 | 1.91 | 62.10 | 60.05 | 60.31 | 2.11 |
| **OS-OCELM_Thr2(Sig)** | 85.23 | 86.38 | 86.25 | 2.20 | 59.19 | 59.79 | 59.81 | 1.75 |
| **OS-AAELM_Thr1(Sig)** | 90.86 | 91.87 | 91.66 | 1.98 | 64.99 | **65.66** | 65.67 | 1.25 |
| **OS-AAELM_Thr2(Sig)** | 89.75 | 90.98 | 90.75 | 2.32 | 61.71 | 64.96 | 64.75 | 2.02 |
| **OS-AAELM_Thr3(Sig)** | 76.01 | 69.61 | 70.33 | 3.63 | 62.97 | 53.11 | 54.30 | 1.43 |

**Table 2(c): Performance of all classifiers on Ecoli and Liver datasets over average of 20 runs**

| | Ecoli | | | | Liver | | | |
|---|---|---|---|---|---|---|---|---|
| | F1 | ACC | AUC | Std_AUC | F1 | ACC | AUC | Std_AUC |
| **knndd** | 62.89 | 90.56 | **89.87** | 2.09 | 41.91 | 33.24 | 51.73 | 1.50 |
| **svdd** | 44.20 | 85.66 | 72.52 | 5.87 | 40.62 | 38.93 | 51.96 | 1.90 |
| **kmeans_dd** | 60.50 | 88.52 | 88.84 | 2.44 | 41.60 | 35.04 | 51.83 | 1.39 |
| **parzen_dd** | 62.14 | 89.52 | 89.65 | 2.58 | 41.19 | 33.42 | 50.97 | 1.87 |
| **nparzen_dd** | 37.25 | 76.15 | 79.47 | 3.82 | 41.66 | 33.53 | 51.53 | 1.71 |
| **pca_dd** | 21.93 | 56.42 | 62.59 | 6.22 | 41.93 | 36.89 | 52.67 | 1.49 |
| **mpm_dd** | 42.67 | 83.31 | 76.04 | 5.45 | 23.16 | **72.78** | 54.51 | 2.26 |
| **incsvdd** | 39.18 | 72.90 | 76.39 | 13.43 | 41.33 | 32.17 | 50.79 | 1.63 |
| **som_dd** | NAN | 91.63 | 56.47 | 5.61 | 39.91 | 47.46 | 53.41 | 2.07 |
| **autoenc_dd** | 46.61 | 82.79 | 80.39 | 6.81 | 42.26 | 38.90 | 53.52 | 2.51 |
| **gauss_dd** | 66.94 | 92.39 | 88.68 | 2.62 | 41.97 | 33.33 | 51.84 | 1.14 |
| **OCELM_Thr1** | 41.80 | 80.61 | 75.88 | 7.60 | 40.83 | 40.63 | 52.38 | 2.59 |
| **OCKELM_Thr1** | 77.04 | 96.35 | 86.04 | 4.13 | 42.18 | 42.06 | 54.16 | 2.03 |
| **Proposed Methods** | | | | | | | | |
| **OCKELM_Thr2** | **77.77** | **96.47** | 86.98 | 6.06 | 40.05 | 59.19 | 56.81 | 2.68 |
| **AAKELM_Thr1** | 20.31 | 40.82 | 63.25 | 3.50 | 41.21 | 38.07 | 52.24 | 1.59 |
| **AAKELM_Thr2** | 65.19 | 93.65 | 81.86 | 6.65 | 40.52 | 41.93 | 52.56 | 2.66 |
| **AAKELM_Thr3** | 56.98 | 90.15 | 83.00 | 3.50 | **42.36** | 44.21 | 54.86 | 1.62 |
| **OCELM_Thr2** | 40.05 | 80.39 | 76.89 | 5.66 | 39.94 | 45.97 | 53.04 | 3.51 |
| **AAELM_Thr1** | 61.73 | 90.37 | **88.37** | 3.13 | 41.54 | 47.37 | 54.92 | 2.27 |
| **AAELM_Thr2** | 75.28 | 96.02 | 85.33 | 3.82 | 41.37 | 57.78 | **57.35** | 2.64 |
| **AAELM_Thr3** | 15.48 | 8.40 | 50.01 | 0.04 | 41.82 | 26.80 | 50.03 | 0.64 |
| **OS-OCELM_Thr1(RBF)** | 34.68 | 74.35 | 73.77 | 5.32 | 41.68 | 48.66 | 55.31 | 2.57 |
| **OS-OCELM_Thr2(RBF)** | 35.01 | 80.32 | 71.44 | 8.80 | 40.33 | 51.73 | 55.20 | 3.07 |
| **OS-AAELM_Thr1(RBF)** | 50.83 | 85.47 | 80.54 | 5.37 | 38.92 | 55.64 | 55.30 | 3.32 |
| **OS-AAELM_Thr2(RBF)** | 56.36 | 93.58 | 73.87 | 5.95 | 40.97 | 54.39 | 56.28 | 2.21 |
| **OS-AAELM_Thr3(RBF)** | 54.01 | 90.58 | 78.09 | 5.67 | 38.16 | 57.19 | 55.47 | 2.50 |
| **OS-OCELM_Thr1(Sig)** | 32.70 | 73.84 | 74.02 | 5.86 | 42.30 | 43.86 | 54.71 | 3.09 |
| **OS-OCELM_Thr2(Sig)** | 34.46 | 81.50 | 69.11 | 7.10 | 39.20 | 50.99 | 53.79 | 3.99 |
| **OS-AAELM_Thr1(Sig)** | 45.58 | 85.08 | 77.10 | 5.50 | 42.04 | 42.63 | 54.17 | 2.16 |
| **OS-AAELM_Thr2(Sig)** | 69.39 | 94.15 | 86.85 | 5.73 | 41.80 | 42.61 | 53.89 | 3.86 |
| **OS-AAELM_Thr3(Sig)** | 16.21 | 12.48 | 52.24 | 4.29 | NAN | **69.39** | 50.68 | 1.72 |

**Table 2(d): Performance of all classifiers on Sonar and Spectf datasets over average of 20 runs**

|  | Sonar | | | | Spectf | | | |
|---|---|---|---|---|---|---|---|---|
|  | **F1** | **ACC** | **AUC** | **Std_AUC** | **F1** | **ACC** | **AUC** | **Std_AUC** |
| **knndd** | 49.21 | 33.78 | 45.72 | 2.65 | 44.31 | 64.80 | 74.98 | 2.17 |
| **svdd** | 23.40 | 67.73 | 56.08 | 2.36 | 43.56 | 79.10 | 68.20 | 3.90 |
| **kmeans_dd** | 50.80 | 57.80 | 58.37 | 4.70 | 51.59 | 78.65 | 76.34 | 3.74 |
| **parzen_dd** | NAN | 63.82 | 50.00 | 0.00 | 54.42 | **90.28** | 68.88 | 3.73 |
| **nparzen_dd** | 49.08 | 33.06 | 45.23 | 2.09 | 32.00 | 40.75 | 60.56 | 0.89 |
| **pca_dd** | 50.95 | 59.90 | 59.45 | 2.35 | 54.12 | 85.40 | 73.40 | 5.24 |
| **mpm_dd** | NAN | 63.82 | 50.00 | 0.00 | 53.50 | 90.18 | 68.56 | 4.66 |
| **incsvdd** | 57.89 | 55.59 | 61.84 | 2.79 | 46.20 | 68.12 | 75.87 | 1.88 |
| **som_dd** | 54.18 | 65.69 | 63.59 | 4.77 | 53.65 | 89.25 | 69.40 | 3.87 |
| **autoenc_dd** | 48.23 | 38.65 | 47.41 | 3.63 | 50.40 | 76.58 | 76.37 | 3.15 |
| **gauss_dd** | NAN | 63.82 | 50.00 | 0.00 | 51.98 | 89.93 | 67.77 | 3.86 |
| **OCELM_Thr1** | 52.88 | 57.89 | 59.57 | 3.97 | 41.33 | 63.52 | 71.02 | 2.63 |
| **OCKELM_Thr1** | 56.44 | 64.18 | 64.16 | 3.06 | 53.75 | 80.35 | 77.48 | 3.17 |
| **Proposed Methods** | | | | | | | | |
| **OCKELM_Thr2** | 57.18 | 75.13 | 68.86 | 2.81 | 52.51 | 80.38 | 76.19 | 3.88 |
| **AAKELM_Thr1** | 49.16 | 36.22 | 46.84 | 1.83 | **59.24** | 87.38 | 75.87 | 3.21 |
| **AAKELM_Thr2** | 40.03 | 39.18 | 42.98 | 2.76 | 49.76 | 80.18 | 73.30 | 3.72 |
| **AAKELM_Thr3** | 62.02 | **75.89** | 71.33 | 3.62 | 57.88 | 80.80 | **82.42** | 3.39 |
| **OCELM_Thr2** | 49.99 | 62.24 | 60.12 | 4.05 | 40.14 | 71.38 | 67.26 | 4.37 |
| **AAELM_Thr1** | 61.55 | 70.59 | 69.46 | 3.10 | 50.15 | 74.40 | 77.59 | 2.28 |
| **AAELM_Thr2** | 53.01 | 72.70 | 66.27 | 3.38 | 49.20 | 81.74 | 71.93 | 4.70 |
| **AAELM_Thr3** | 65.26 | 66.71 | 70.89 | 3.11 | 30.12 | 26.33 | 55.48 | 7.46 |
| **OS-OCELM_Thr1(RBF)** | 54.50 | 61.25 | 62.02 | 4.33 | 30.91 | 52.87 | 57.25 | 7.54 |
| **OS-OCELM_Thr2(RBF)** | 49.81 | 65.46 | 61.66 | 4.10 | 30.42 | 60.78 | 57.34 | 6.28 |
| **OS-AAELM_Thr1(RBF)** | 56.11 | 66.28 | 64.80 | 4.29 | 32.30 | 48.55 | 57.55 | 8.42 |
| **OS-AAELM_Thr2(RBF)** | 54.13 | 67.34 | 64.47 | 4.65 | 27.12 | 56.51 | 54.38 | 6.12 |
| **OS-AAELM_Thr3(RBF)** | 39.91 | 68.42 | 60.89 | 6.98 | NAN | 84.57 | 50.59 | 1.36 |
| **OS-OCELM_Thr1(Sig)** | 54.50 | 54.18 | 58.82 | 5.39 | 42.54 | 77.69 | 67.14 | 4.66 |
| **OS-OCELM_Thr2(Sig)** | 52.35 | 56.45 | 58.55 | 5.49 | 36.10 | 76.45 | 62.50 | 5.62 |
| **OS-AAELM_Thr1(Sig)** | 54.64 | 70.66 | 66.11 | 3.62 | 48.92 | 75.35 | 75.29 | 3.26 |
| **OS-AAELM_Thr2(Sig)** | 55.11 | 71.48 | 66.55 | 4.42 | 48.06 | 79.27 | 72.28 | 4.66 |
| **OS-AAELM_Thr3(Sig)** | **65.63** | 72.70 | **72.59** | 4.01 | 38.60 | **88.17** | 62.13 | 3.26 |

As our proposed methods heavily depend on three factors viz., threshold criteria, regularization parameter and kernel parameter so, **Fig. 9(a)-(e)** depict the behavior of the proposed methods viz., **AAKELM** and **OCKELM**, over changing the kernel parameters (KERN_PAR), regularization parameter (C) and threshold criteria. **Fig. 9(a)-(e)** is depicted for RBF kernel; however we tested with four kernels viz., RBF, Polynomial, linear and Wavelet. As you can see in **Fig. 9(a)-(e)**, behaviour of the proposed methods for **Thr1** and **Thr2** does not change much but it performs totally different for **Thr3**. However, performance of **Thr3** is comparable to other threshold criteria viz., **Thr1** and **Thr2**. **Thr1** performed overall better compared to other two threshold criteria but **Thr3** also performed well, however there is basic difference between applicability of both the threshold criteria. If you have enough amounts of data and some data among target data distributed differently from remaining dataset or they are at boundary point of distribution. As we know that **Thr1** is decided by rejecting some of the data from the training set then there is always a possibility that testing set data belongs to the rejected part of the data and it can heavily impact the performance of the classifier whereas **Thr3** will perform better in that case because **Thr3** criterion does not reject any amount of data so it will also not reject any amount of data during testing.

**Following points to be noted regarding proposed classifiers:**

(i) Among all ELM based OCC methods, reconstruction based OCC methods i.e. autoencoder based OCC methods performed best for all eight datasets. So, our proposed methods outperformed methods proposed by Leng et al. [28].

(ii) Except Breast Cancer and Ecoli dataset, our proposed methods outperformed all existing methods for six out of eight datasets and also exhibited slightly less AUC value compared to rest of the two datasets viz., Breast Cancer and Ecoli dataset.

(iii) In online sequential version of OCC methods, additive hidden nodes performed better in case of five datasets and RBF hidden nodes performed better for three datasets. In this place also, reconstruction based i.e. autoencoder based online sequential OCC method performed better than boundary based online sequential OCC methods for all 8 datasets.

(iv) Among all offline version of ELM based OCC methods, kernel feature based OCC methods performed best for five datasets and random feature based methods performed best for remaining three datasets.

(v) Among regularization coefficient (**C**) and kernel parameter (**kern_par** or $\sigma$), we kept $\sigma$ on higher priority compared to **C** i.e. when we obtained the consistent boundary using different parameter combinations, ($\sigma_1$, **C₁**) and ($\sigma_2$, **C₂**) then we preferred smaller $\sigma$ over larger **C**. Every possible combination of $\sigma$ and **C** have been employed to obtain most complex classifier as long as classifier is consistent. Range of **C**: [$10^{-8}$, $10^{-8}$........., $10^7$, $10^8$] and range of $\sigma$: twenty values are selected between minimum and maximum pair wise distance among records of the dataset.

(vi) As we have discussed that, **Thr3** criterion doesn't reject any percent of data, so you can see in **Fig. 7** and **8**, it covers all samples during creating boundary but other two thresholds doesn't do the same.

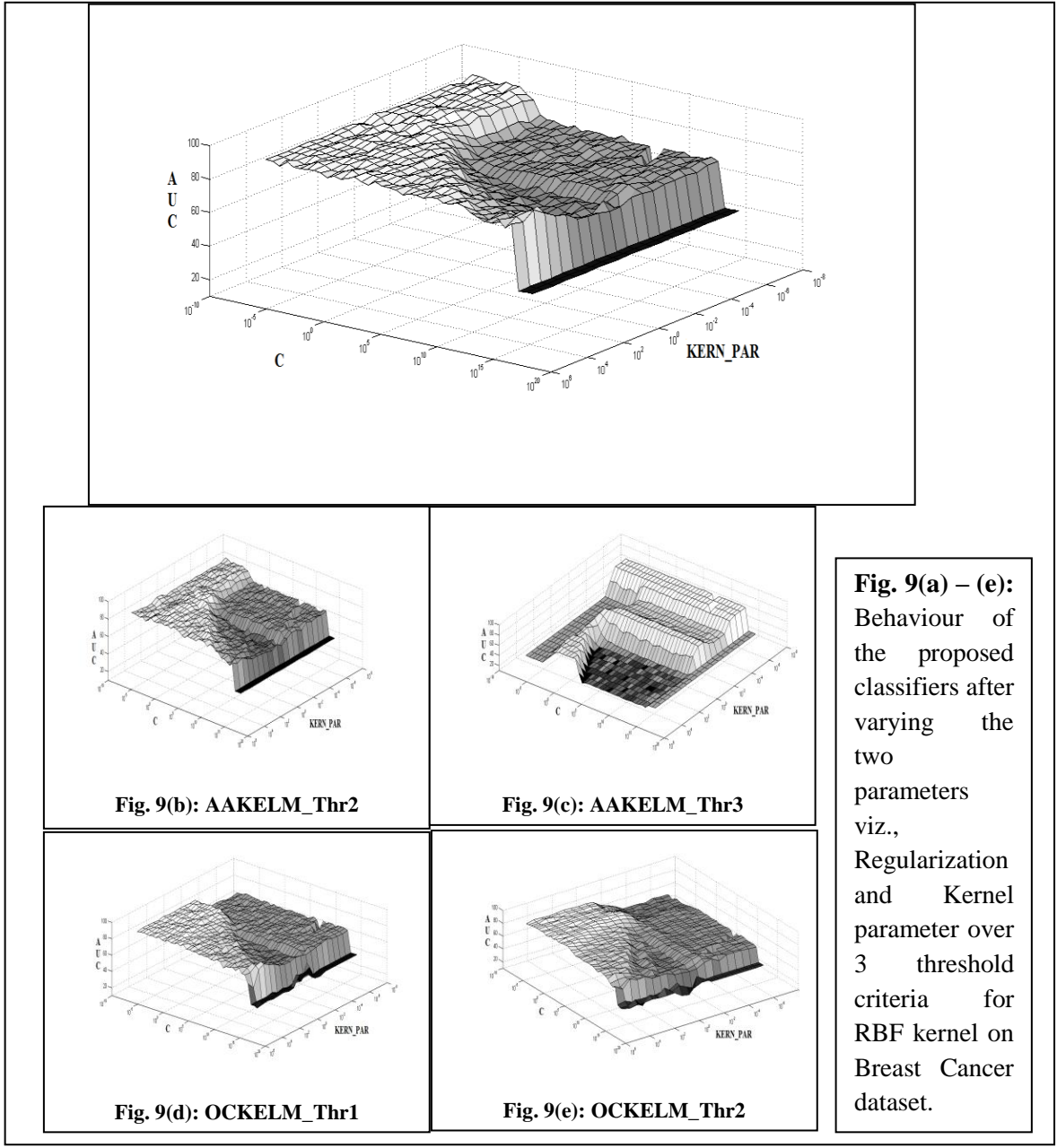

**Fig. 9(b): AAKELM_Thr2**
**Fig. 9(c): AAKELM_Thr3**
**Fig. 9(d): OCKELM_Thr1**
**Fig. 9(e): OCKELM_Thr2**

**Fig. 9(a) – (e):** Behaviour of the proposed classifiers after varying the two parameters viz., Regularization and Kernel parameter over 3 threshold criteria for RBF kernel on Breast Cancer dataset.

## 8.3 Time Comparison Analysis on Benchmark Datasets

As we have discussed in the paper several times that ELM based one-class classifier performed faster compared to traditional methods, so now we are exhibiting time comparison analysis in **Table 3**. We consider two most popular algorithms of OCC viz. **svdd** and **autoenc_dd** for this analysis. As per result presented by Leng et el. [28], testing time does not differ much from ELM based one-class classifier, so we presented only training time of all classifiers. Further, there is no point to compare online sequential algorithm from offline algorithm in term of time because it is obvious that online sequential will take more time due to sequential acceptance of data. Computational complexity of autoencoder based classifier is more compared to other non-autoencoder (like **OCELM**, **OCKELM**) based classifier because computational complexity autoencoder depends upon both number of samples and dimension of samples.

**Table 3: Training Time (in seconds) of all ELM based, autoencoder based and SVDD based one-class classifiers on all datasets over 20 runs**

|  | Abalone | Arrhythmia | Breast cancer | Diabetese | Ecoli | Liver | Sonar | Spectf heart |
|---|---|---|---|---|---|---|---|---|
| **Svdd** | 31.36 | 0.41 | 0.76 | 2.26 | 0.27 | 0.57 | 0.32 | 0.37 |
| **autoenc_dd** | 58.14 | 732.04 | 12.22 | 32.53 | 4.08 | 5.12 | 12.32 | 16.67 |
| **OCKELM_Thr1** | 0.48 | **0.07** | **0.06** | **0.10** | **0.05** | **0.06** | **0.06** | **0.06** |
| **OCKELM_Thr2** | 0.50 | **0.07** | 0.07 | **0.10** | 0.06 | **0.06** | **0.06** | **0.06** |
| **AAKELM_Thr1** | **0.49** | 0.08 | 0.07 | 0.16 | **0.05** | **0.06** | **0.06** | **0.06** |
| **AAKELM_Thr2** | 0.52 | 0.08 | 0.07 | 0.14 | 0.06 | **0.06** | **0.06** | **0.06** |
| **AAKELM_Thr3** | 0.57 | 0.16 | 0.07 | 0.11 | 0.05 | 0.07 | **0.06** | **0.06** |
| **OCELM_Thr1** | 0.51 | 0.15 | 0.09 | 0.14 | 0.06 | 0.07 | 0.07 | 0.07 |
| **OCELM_Thr2** | 0.54 | 0.15 | 0.09 | 0.14 | 0.06 | 0.08 | 0.07 | 0.08 |
| **AAELM_Thr1** | 0.56 | 0.16 | 0.09 | 0.13 | 0.06 | 0.07 | 0.09 | 0.08 |
| **AAELM_Thr2** | 0.56 | 0.16 | 0.09 | 0.14 | 0.06 | 0.08 | 0.08 | 0.08 |
| **AAELM_Thr3** | 0.57 | 0.23 | 0.09 | 0.17 | 0.06 | 0.07 | 0.08 | 0.08 |

Whereas non-autoencoder based classifiers depend upon only number of samples. Reason behind inferior performance of **autoenc_dd** and **svdd** is that traditional autoencoder needs to tune the parameters and svdd needs to solve a quadratic programming problem to classify the data but ELM based classifiers do not need to tune the parameters to classify the data.

## 9. Conclusion and Future Direction

This paper delineates eight offline and five online methods for OCC and exhibits the effectiveness on the two artificial datasets and eight benchmark datasets from different disciplines. Autoencoder has been successfully employed earlier for OCC but it needs to tune the parameters but our proposed autoencoders for OCC yield results in just one pass without tuning any parameters. Our proposed online OCC approaches are pretty fast and simple compared to earlier OCC methods which are very time consuming and cumbersome. Methods presented in this paper outperformed the existing traditional method based one-class classifiers as well as existing ELM based one-class classifiers. This paper **expanded the DD Toolbox** proposed in [60] by incorporating ELM based one-class classifiers with this and provided significant contribution[#]. We can use our toolbox for other ELM-variants also by just simple modification in the code of the presented toolbox. In optimal parameter search by consistency based model selection, there is a possibility that it can stuck in local minima. This is due to the fact that our model selection method perform linear search over various combination of values of parameters during optimal parameter selection, so we need some other optimization techniques like Genetic Algorithm (GA), Particle Swarm Optimization (PSO) etc. which can utilize whole search space for optimal selection of parameter. In consistency based model selection, if you have to optimize only one or two parameters then it performs pretty fast but if you will endeavor to optimize more than two parameters then it exhibits slow convergence. This fact is quite obvious with our kernelized based approaches which exhibits very fast convergence in case of RBF kernel because RBF kernel have only one parameter to optimize but exhibits slow convergence in case of Wavelet kernel because Wavelet kernel has three parameters to optimize.

This issue can be addressed by some multidimensional optimization algorithm like multidimensional PSO. However, our methods shows fast convergence and better accuracy compared to existing methods. So, we conclude that our proposed methods can be a viable and effective alternative for OCC and presented toolbox would be very helpful in further research on OCC. Possible future directions of our work are to enhance it to multilayer network, design it for regularized and kernelized version of online sequential version and as the current challenges need, to make it scalable for big data handling etc.

#**Our expanded toolbox will be publicly available after acceptance of the paper on this link: http://cgautam.wix.com/chandan#!research-blog/cswb**